\documentclass[10pt,twocolumn,letterpaper]{article}

\usepackage{authblk}
\usepackage{times}
\usepackage{epsfig}
\usepackage{graphicx}
\usepackage{amsmath}
\usepackage{amssymb}
\usepackage{booktabs}

\usepackage{todonotes}

\usepackage{microtype}
\usepackage{graphicx}
\usepackage{subfigure}
\usepackage{booktabs} 
\usepackage{multirow}
\usepackage{csquotes}
\usepackage{threeparttable} 
\usepackage{enumitem} 

\newcommand{\eat}[1]{}

\usepackage[breaklinks=true,bookmarks=false]{hyperref}

\date{\vspace{-5ex}}

\begin{document}

\title{Point-Syn2Real: Semi-Supervised Synthetic-to-Real\\ Cross-Domain Learning
for Object Classification in 3D Point Clouds}

\author[1]{Ziwei Wang}
\author[1]{Reza Arablouei}
\author[1]{Jiajun Liu}
\author[1]{Paulo Borges}
\author[1]{\\Greg Bishop-Hurley}
\author[2]{Nicholas Heaney \thanks{major part of this work was done while N. Heaney was at CSIRO}}
\affil[1]{The Commonwealth Scientific and Industrial Research Organisation}
\affil[2]{Evolve Group}
\affil[ ]{\textit {\{ziwei.wang, reza.arablouei, jiajun.liu, paulo.borges, greg.bishop-hurley\}@csiro.au}}
\affil[ ]{\textit {nicheaney@me.com}}

\maketitle
\thispagestyle{empty}

\begin{abstract}
Object classification using LiDAR 3D point cloud data is critical for modern applications such as autonomous driving.
However, labeling point cloud data is labor-intensive as it requires human annotators to visualize and inspect the 3D data from different perspectives.
In this paper, we propose a semi-supervised cross-domain learning approach that does not rely on manual annotations of point clouds and performs similar to fully-supervised approaches. We utilize available 3D object models to train classifiers that can generalize to real-world point clouds.
We simulate the acquisition of point clouds by sampling 3D object models from multiple viewpoints and with arbitrary partial occlusions. We then augment the resulting set of point clouds through random rotations and adding Gaussian noise to better emulate the real-world scenarios.
We then train point cloud encoding models, e.g., DGCNN, PointNet++, on the synthesized and augmented datasets and evaluate their cross-domain classification performance on corresponding real-world datasets.
We also introduce Point-Syn2Real, a new benchmark dataset for cross-domain learning on point clouds.
The results of our extensive experiments with this dataset demonstrate that the proposed cross-domain learning approach for point clouds outperforms the related baseline and state-of-the-art approaches in both indoor and outdoor settings in terms of cross-domain generalizability.
The code and data will be available upon publishing.
\end{abstract}


\section{Introduction} \label{sec:introduction}

%


\begin{figure}[t]
    \centering
    \includegraphics[width=0.47\textwidth]{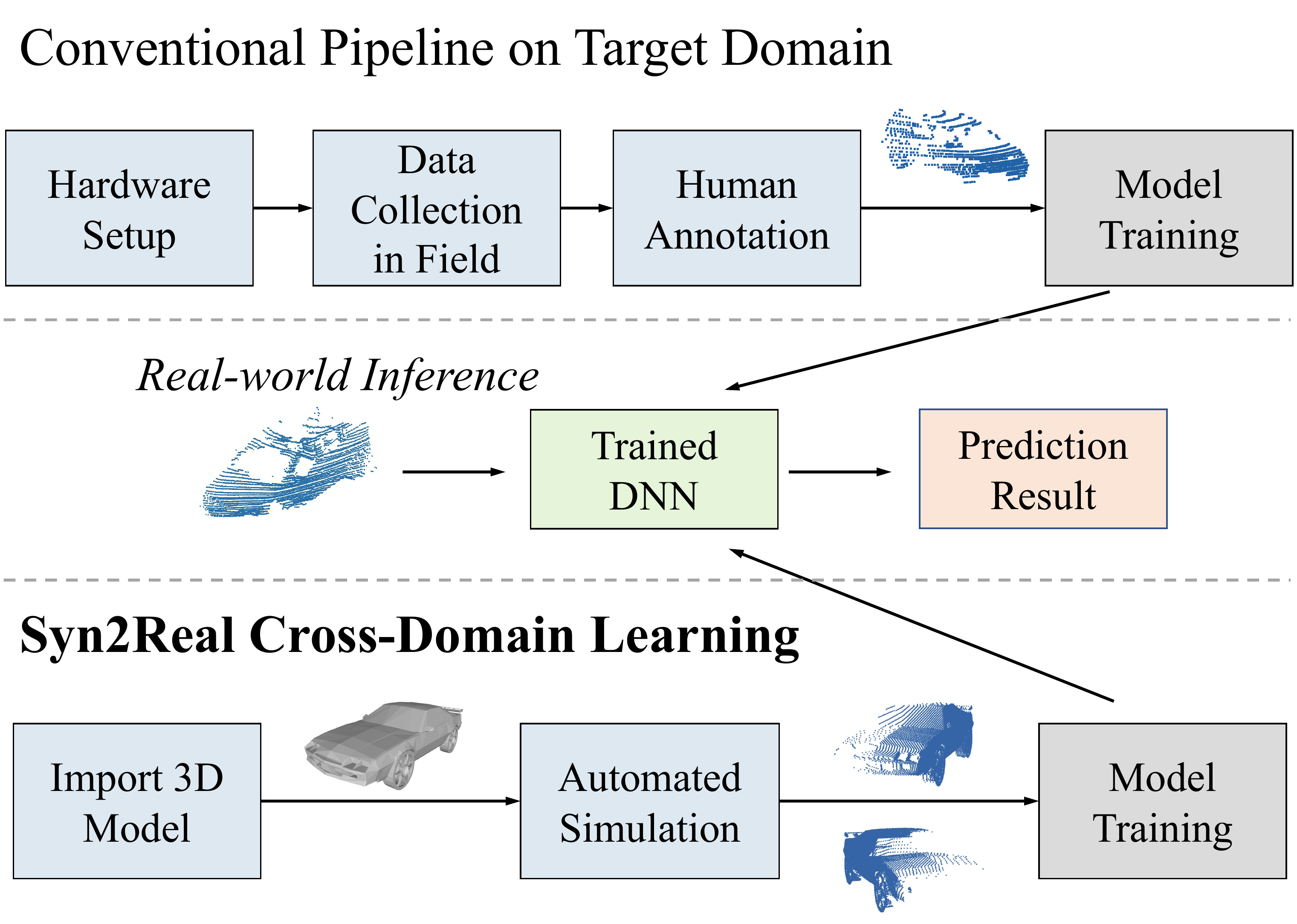}
    \caption{An illustration of single-domain learning versus synthetic-to-real cross-domain learning. The single-domain approach requires collecting real-world point cloud data and annotating it. The cross-domain approach trains the model on the data synthesized through automated simulation and utilizes the learned model for inference on real data.}
    \label{fig:intro_example}
\end{figure}

Detecting and identifying the objects present in a scene is a crucial but challenging task in machine learning.
Substantial efforts have been made to develop algorithms that can accurately recognize objects~\cite{sift,resnet}, detect objects~\cite{yolov4,swin_transformer2}, track objects, and recognize human-object interactions~\cite{devnet} using images or videos captured by cameras. 
Although object detection in computer vision has significantly advanced in recent years,
there exist fundamental limitations. For example, most cameras have difficulty capturing clear images in low- or excessive-light conditions, or determining the exact distance of objects in 2D images is challenging. In addition, privacy concerns often arise around images as they may contain private information easily perceivable by humans.

Light detection and ranging (LiDAR) sensors use laser beams to scan their surroundings and construct 3D representations of the objects within. 
The scanned 3D snapshots are stored as the so-called point clouds. Detecting objects from 3D point clouds can help resolve some of the challenges associated with image-based object detection. 
A LiDAR scanner can obtain precise information of object positions and shapes regardless of lighting conditions. Moreover, as point clouds are less perceivable by humans, they can help enhance privacy preservation.
Given the above advantages, computer vision applications can benefit from 3D point clouds.
For example, in autonomous driving, when the visibility is poor, LiDAR sensors can help detect obstacles.



Recognizing objects in point clouds using machine learning has been studied by many researchers.
Conventional approaches~\cite{fpfh, iss_keypoints} use carefully designed features to represent various shapes in point clouds.
More recently, deep learning (DL) models have been used to learn point-level and object-level features in an end-to-end manner. A notable example is PointNet~\cite{pointnet17} that uses end-to-end DL. It has led to significant improvement in point cloud classification and segmentation performance. Ensuing methods, including PointNet++~\cite{pointnet2} and DGCNN~\cite{dgcnn}, consider local neighborhood information for refined feature extraction.
Nonetheless, training DL models that perform well on real-world data is challenging. Firstly, DL-based methods require large amounts of labeled data for training, whose acquisition is slow and laborious. Secondly, the sensed back-scatter laser light in LiDAR scans is inevitably corrupted by noise that can affect the performance of the learned model. Third, real-world point cloud data is often subject to partial occlusions that can also affect the performance of the learned model.


Databases of models created for 3D graphics design contain large collections of high-quality synthetic 3D models of various known objects. These databases can be used to train new machine learning models to recognize objects in real-world applications within complex environments while minimizing the need for human annotations of point cloud data.
In this paper, we propose a novel synthetic-to-real cross-domain learning approach for point cloud data, called Point-Syn2Real. It enables learning end-to-end DL-based models for classifying objects in real-world point clouds by making use of the available synthetic 3D model databases.
With Point-Syn2Real, we learn classification models from synthetic 3D object data (the source domain) and extend the knowledge gained from the synthetic data to real-world point cloud data (the target domain) as illustrated in Figure~\ref{fig:intro_example}. Since there are substantial discrepancies between the characteristics of the synthetic and real data, an object classification model trained on the source domain does not usually perform well on the target domain, when applied directly. Therefore, we pay special attention to improving the generalizability of the learned model.


In the training phase of Point-Syn2Real, we first simulate 3D LiDAR scans for each considered object from multiple viewpoints to generate synthetic but realistic point cloud data. We also emulate arbitrary partial occlusions that can occur in real-world 3D scans. We then apply several random rotations and add Gaussian noise to the generated synthetic data. Augmenting the datasets via rotation and noise addition helps the trained models learn feature representations that are rotation-invariant and robust to noise. We feed the simulated and augmented synthetic 3D point cloud data into a DL-based point-cloud feature encoder, e.g., DGCNN or PointNet++, and aggregate the learned point features via max-pooling to preserve the most salient features. We then pass the features through a multilayer perceptron (MLP) classifier to generate class-wise predictions. We compute the loss function associated to each labeled point cloud as the cross-entropy of predictions and the corresponding ground truth label.
In addition, we utilize the unlabeled real-world point cloud data of the target domain for training in a semi-supervised learning fashion via
entropy minimization~\cite{entropy_loss_nips04}. To this end, we feed the unlabeled point clouds through the feature extractor and object classifier to create the respective class-wise predictions. The loss function for each unlabeled point cloud is the entropy of its corresponding predicted probabilities. This encourages the learned model to make more confident predictions on the unlabeled data and consequently improves its generalization ability.

During inference on real-world data of the target domain, the learned model encodes the input point cloud data into object-level features, and the classifier predicts the associated object class based on the features.
To evaluate the effectiveness of the proposed Point-Syn2Real approach, we conduct experiments with both indoor and outdoor object classification datasets, and present the results using various performance metrics. 


Our key contributions in this paper are:
\setlist{nolistsep}
\begin{enumerate}[noitemsep]
    \item We introduce a novel semi-supervised cross-domain learning approach that can generalize the knowledge learned from synthetic 3D point clouds to real-world data collected by LiDAR scanners. We use random rotation/noise addition augmentation, multi-view simulation, and entropy minimization to enhance the robustness and performance of the learned models.
    \item We create a comprehensive synthetic-to-real cross-domain 3D point cloud dataset as a benchmark, which includes indoor and outdoor scenarios.
    \item We provide the results of extensive experiments using data of both indoor and outdoor settings, and demonstrate the effectiveness of the proposed approach.
\end{enumerate}


\section{Related Work} \label{sec:related_work}



PointNet~\cite{pointnet17} is one of the first DL-based end-to-end models that can directly process raw point cloud data. It calculates point-level features, which can be aggregated via max-pooling to produce global features. PointNet delivers promising results in point cloud classification and segmentation tasks. However, its functionality is limited as it only considers global features and pays less attention to local geometric features. As an improvement to PointNet, PointNet++~\cite{pointnet2} introduces additional sampling and grouping layers to leverage local information.
DGCNN~\cite{dgcnn} is another DL-based point cloud encoder that builds a nearest-neighbor graph to incorporate the local and global geometric information.
These models exhibit good performance when trained and evaluated on data from the same domain. However, they usually do not preform well when they are trained on data from one domain and evaluated on data from another domain.
Therefore, models trained on synthetic 3D datasets such as ModelNet~\cite{data_modelnet} and ShapeNet~\cite{data_shapenet} may not perform well on real-world datasets such as ScanNet~\cite{data_scannet}.

Obtaining labeled real-world point cloud data for training is challenging due to the associated labor costs or time constraints. Cross-domain learning utilizes readily-available synthetic data (source domain) for training and adapts the learned model to perform inference on real-world data (target domain) with limited annotations.
There are a few existing works that address cross-domain learning with point cloud data. 
PointDAN~\cite{pointdan_nips19} aligns the local and global features to mitigate distribution shift between the source and target domains.
DefRec~\cite{defrec_pointda_wacv19} learns a representation model by reconstructing point clouds with induced deformations.
These methods are useful for adapting models from a data distribution perspective. However, they do not explicitly address challenges involved in real-world data acquisition such as partial occlusions and viewpoint variations.
Other cross-domain learning methods such as~\cite{xmuda, cross-sensor, birdview} deal with point-cloud-related tasks other than object classification.


\section{Proposed Approach} \label{sec:method}

\begin{figure}[t]
    \centering
    \includegraphics[width=0.47\textwidth]{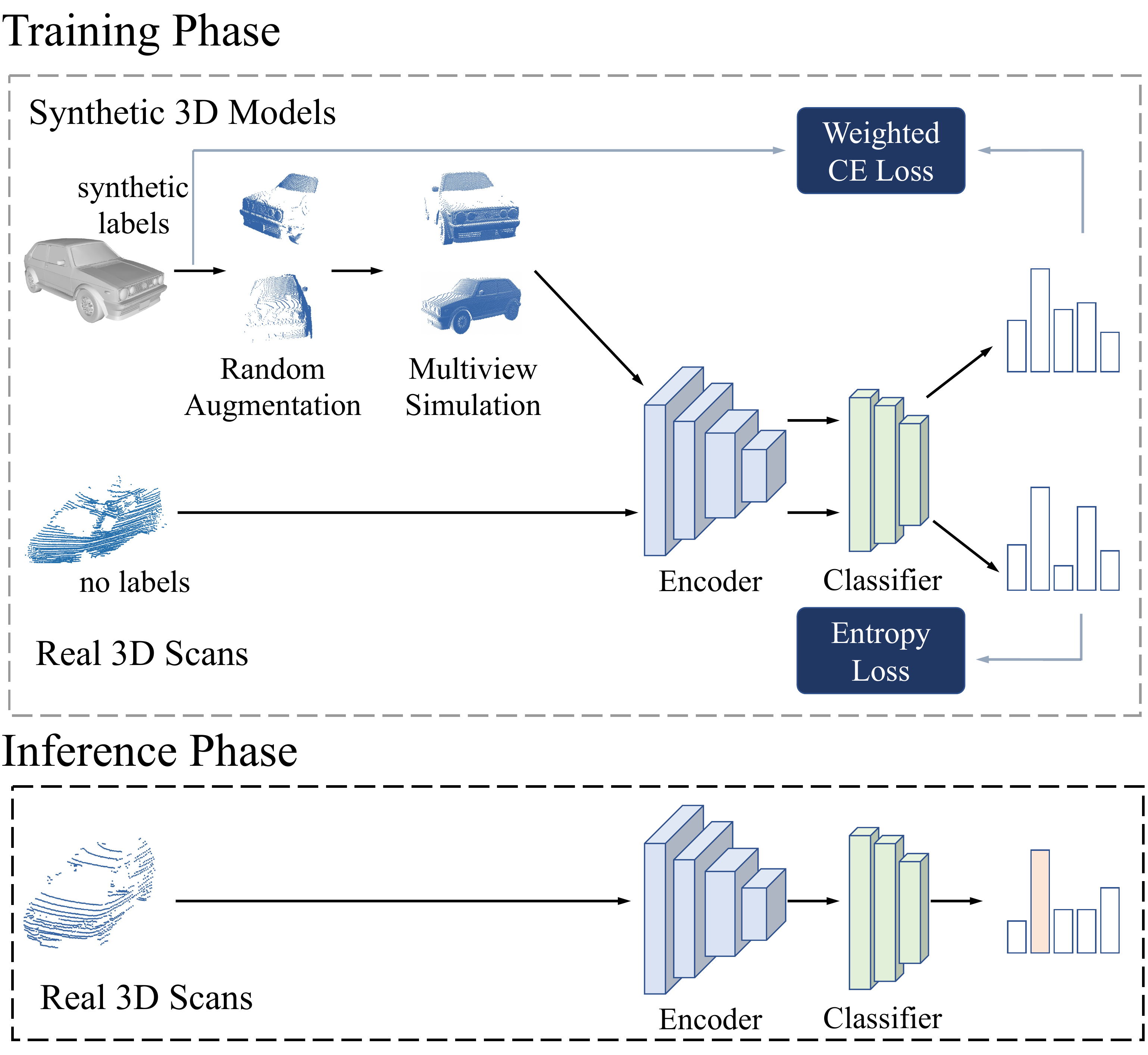}
    \caption{Overview of the proposed synthetic-to-real cross-domain learning approach.}
    \label{fig:method_overview}
\end{figure}

In this section, we describe our proposed semi-supervised approach to synthetic-to-real cross-domain learning. After giving an overview, we explain the procedures of multi-view simulation and data augmentation as well as the utilized point cloud encoder. We also explain the model learning process and the associated objective function designed for semi-supervised learning while coping with class imbalance.

\subsection{Overview}

We provide a visual overview of the proposed approach in Figure~\ref{fig:method_overview}. During training, we use a 3D computer-aided design (CAD) tool to generate a set of partially-occluded point clouds taken from multiple viewpoints for each considered object. We denote the generated point cloud set associated with the $i$th object as $\mathcal{P}_i = \{P_{i1}, P_{i2}, \cdots, P_{iM}\}$ where $M$ is the number of viewpoints. Each point cloud is a set of points in the Euclidean space, i.e., $P_{ij} = \{p_{ij1}, p_{ij2}, \cdots, p_{ijN_{ij}}\}$ where $N_{ij}$ is the number of points in $P_{ij}$ and each point $p_{ijk}$ has three coordinate values $(x_{ijk}, y_{ijk}, z_{ijk})$.
We utilize unlabeled real-world point clouds from the target domain to realize semi-supervise learning. Thus, we denote the set of real point clouds associated with the $i$th object as $S_{i}$.

We augment the set of synthetic point clouds by applying multiple random rotations and adding Gaussian noise. We then feed the augmented synthetic data into a DL-based point cloud encoder to extract point-level features.
We aggregate these features via max-pooling before forwarding to fully-connected (FC) classification layers, which output predicted posterior probabilities for each object class using the $\mathrm{softmax}$ function.
To jointly train the neural networks of the encoder and the classifier, we use a composite objective function that aggregates the losses associated with both labeled synthetic point cloud data and unlabeled real point cloud data. For the labeled synthetic data, we use the cross-entropy loss, and, for the unlabeled real data, we use the entropy loss. We weight both losses appropriately to account for the class imbalance.
At inference time, we feed point clouds produced by LiDAR scanners into the relevant trained model to make predictions.

\subsection{Multi-view Point Cloud Simulation}\label{mvs}

In LiDAR scans, the objects of interest may be occluded by other objects or even themselves. In our cross-domain learning approach, we simulate occlusions in synthesizing the training set.
similar to RotationNet~\cite{rotationnet}, which takes snapshots from multiple viewpoints to create multi-view 2D images, one can synthesize multi-view 3D point clouds~\cite{pcn_render, occo_iccv21}. In particular, given a 3D object model, one can simulate realistic LiDAR scans and generate multiple point clouds of the object from different viewpoints.

\begin{figure}[t]
    \centering
    \includegraphics[width=0.47\textwidth]{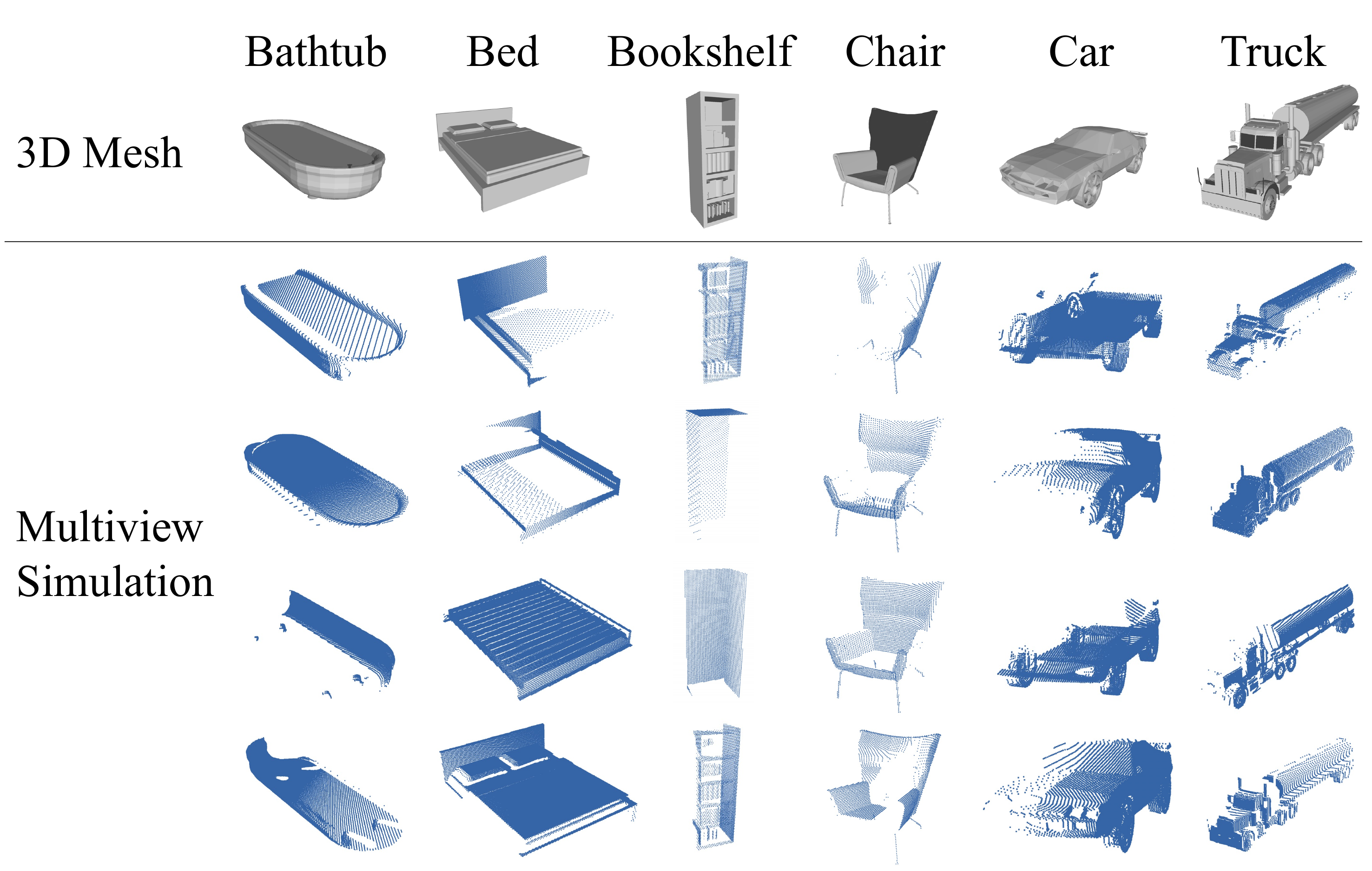}
    \caption{An example synthetic point cloud dataset with six objects and four viewpoints.}
    \label{fig:eval_multiview}
\end{figure}

In this work, we utilize the open-source software Blender\footnote{\url{https://www.blender.org/}} to create synthetic point clouds for training. The procedure has two major steps, i.e., depth map simulation and back projection. The 3D model is positioned in the center of the scene at coordinates $(0,0,0)$. The depth sensor is set up in a random position with an empirically adjusted maximum distance to the object. Its intrinsic and extrinsic properties are recorded for 3D reconstruction.
After setting up the scene, a snapshot of the depth map is captured and saved. A partially-occluded 3D scan is then generated by back-projecting the depth map~\cite{pcn_render}. We select the new position of the depth sensor randomly and repeat the simulation procedure $M$ times for each object to generate the sets of point clouds $\mathcal{P}_i, \forall i\in O$, where $O$ is the set of objects.
In Figure~\ref{fig:eval_multiview}, we illustrate an example synthetic point cloud dataset with six objects and four viewpoints.


\subsection{Data Augmentation}\label{da}

We augment the synthesized partially-occluded point clouds by applying random rotations and adding Gaussian noise to improve the robustness and accuracy of the learned models.

\subsubsection{Random Rotation}

We rotate each point cloud around the $z$-axis by a uniformly-distributed random angle, i.e., $\varphi \in [0,2\pi]$. The rotation of every point of the point cloud $P_{ij}$, i.e., $p_{ijk} = (x_{ijk},y_{ijk},z_{ijk})$, around the $z$-axis by $\varphi$ is expressed via the following linear transformation
\begin{equation}
    \begin{bmatrix}
        x^\prime_{ijk}\\
        y^\prime_{ijk}\\
        z^\prime_{ijk}
    \end{bmatrix}
    =
    \begin{bmatrix}
        \cos(\varphi) & \sin(\varphi) & 0\\
        -\sin(\varphi) & \cos(\varphi) & 0\\
        0 & 0 & 1
    \end{bmatrix}
    \begin{bmatrix}
        x_{ijk}\\
        y_{ijk}\\
        z_{ijk}
    \end{bmatrix}
\end{equation}
We denote the rotated point cloud as $P^\prime_{ij}$.

\subsubsection{Gaussian Noise}
When collecting data in real world, sensing imperfections due to, e.g., measurement noise or error, may corrupt the data. Therefore, to make our synthetic point cloud data more realistic, we add noise to the values of each synthetic point as
\begin{equation}
    p''_{ijk} = p^\prime_{ijk} + \nu_{ijk},
\end{equation}
where $\nu_{ijk}=(\nu_{xijk},\nu_{yijk},\nu_{zijk})$ is the additive noise with $\nu_{cijk}, \forall c\in\{x,y,z\},$ being independently drawn from a Gaussian distribution with mean $\mu=0$ and standard deviation $\sigma=0.01$.

\subsection{Cross-domain Point Cloud Encoder}
In this section, we elaborate the point cloud encoder and classifier, and describe the unified learning objective that consists of cross entropy loss for labeled synthetic data, and entropy loss for unlabeled real data.

Given a partially-occluded and randomly-augmented point cloud $P''_{ij}$, the point cloud encoder, denoted by the function $f(\cdot)$, takes the point cloud as the input and outputs the point-level feature vector of dimension $D$, i.e., $\mathbf{f}_{ij}=f(P''_{ij})$. The point features are then aggregated using an effective symmetric aggregation function, i.e., max-pooling denoted by $\mathrm{max\text{-}pool(\cdot)}$, to produce the pooled global feature vector, i.e., $\mathbf{g}_{ij}=\mathrm{max\text{-}pool}(\mathbf{f}_{ij})\in\mathbb{R}^{D}$.
The global feature vector is then passed through a classifier, $h(\cdot)$, that is a multilayer fully-connected neural network (perceptron) and outputs the logits for each class. The $\mathrm{softmax}(\cdot)$ function is applied to the logits to produce the class-wise posterior probabilities, denoted by $\mathbf{q}_{ij}\in\mathbb{R}^C$ where $C$ is the number of classes, i.e.,
$\mathbf{q}_{ij} = \mathrm{softmax}(h(\mathbf{g}_{ij}))$.
We calculate the categorical cross-entropy loss that evaluates the divergence between the predicted posterior probabilities and the ground-truth label as
\begin{equation}
l_{ij} = -\mathbf{y}_{ij}^\intercal \log(\mathbf{q}_{ij})
\end{equation}
where $\mathbf{y}_{ij}\in\mathbb{R}^{C}$ is the one-hot vector for the ground-truth label corresponding to $P_{ij}$.
We use a weighted version of the cross-entropy loss to mitigate the impact of class imbalance~\cite{focalloss}.

To exploit the information available through the unlabeled real point clouds from the target domain, we utilize the entropy loss function calculated as
\begin{equation}
\ell_{it}=-\mathbf{s}_{it}^\intercal \log(\mathbf{s}_{it}) 
\end{equation}
where $\mathbf{s}_{it}$ is the vector of posterior probabilities predicted by the model for the $t$th real point cloud of the $i$th object available for training. Minimizing the entropy loss for unlabeled data encourages the learned model to make more confident predictions, which can in turn improve its performance. The unified objective function that we minimize during training is the weighted average of the cross-entropy and entropy losses for all available synthetic and real point clouds.

The specifications of the encoder is not central to our approach. Hence, any suitable point cloud encoder can be used. In this work, we primarily use DGCNN~\cite{dgcnn} as the point cloud encoder since it is efficient and leads to good performance.
We also consider using PointNet++~\cite{pointnet2} as our point cloud encoder in Section~\ref{sec:applicability}. The main common property of DGCNN and PointNet++ is their preservation of local neighborhood in feature calculation.
DGCNN constructs a $k$-nearest neighbor (KNN) graph in every graph convolutional layer. The first KNN graph is built upon raw 3D point coordinates to preserve geometric local neighborhood information. In the second and following layers, the local neighborhood is defined in the feature space.
PointNet++ uses a hierarchical point-set feature learning module to recursively sample and group points in local regions.
In both encoders, the computed point-level features are max-pooled to yield global features. 


\section{Point-Syn2Real Dataset} \label{sec:dataset}

\begin{table}[h]
\centering
\footnotesize
\caption{Overview of the Point-Syn2Real dataset.}
\begin{threeparttable}[t]
\begin{tabular}{lccrr}
    \toprule
    Settings                 & Domain   & Name          & \begin{tabular}[c]{@{}c@{}}Original\\Models\end{tabular} & \begin{tabular}[c]{@{}c@{}}Simulated\\Scans\end{tabular} \\
    \midrule
    \multirow{3}{*}{Indoor}  & S\tnote{\dag}      & ModelNet      & 4,183      & 41,830   \\
                             & S                  & ShapeNet      & 17,378     & 173,780   \\
                             & T\tnote{\ddag}     & ScanNet       & 7,879      & --        \\ 
    \midrule
    \multirow{2}{*}{Outdoor} & S      & 3D\_City        & 3,116     & 31,160   \\
                             & T      & SemKitti\_Obj   & 77,908   & --       \\ 
    \bottomrule
\end{tabular}
    \begin{tablenotes}
     \item[\dag] S - Synthetic Source Domain
     \item[\ddag] T - Real-world Target Domain
   \end{tablenotes}
\end{threeparttable}

\label{tab:dataset_overview}
\end{table}

We compile a new benchmark dataset, called Point-Syn2Real, by gathering data from multiple sources. The dataset can be used to evaluate the performance of cross-domain learning methods that involve transferring knowledge from synthetic 3D data to real-world point cloud data.
Point-Syn2Real covers both indoor and outdoor settings as shown in Table~\ref{tab:dataset_overview}. The multiview synthesis considerably increases the number of instances available for training.

For the indoor setting, we extracted ten overlapping categories from ModelNet~\cite{data_modelnet}, ShapeNet~\cite{data_shapenet}, and ScanNet~\cite{data_scannet} datasets following the protocols described in \cite{pointdan_nips19}. In particular, ModelNet and ShapeNet constitute the source domain of synthetic data  and ScanNet forms the target domain of real-world data.

\begin{figure}[b]
    \centering
    \includegraphics[width=0.44\textwidth]{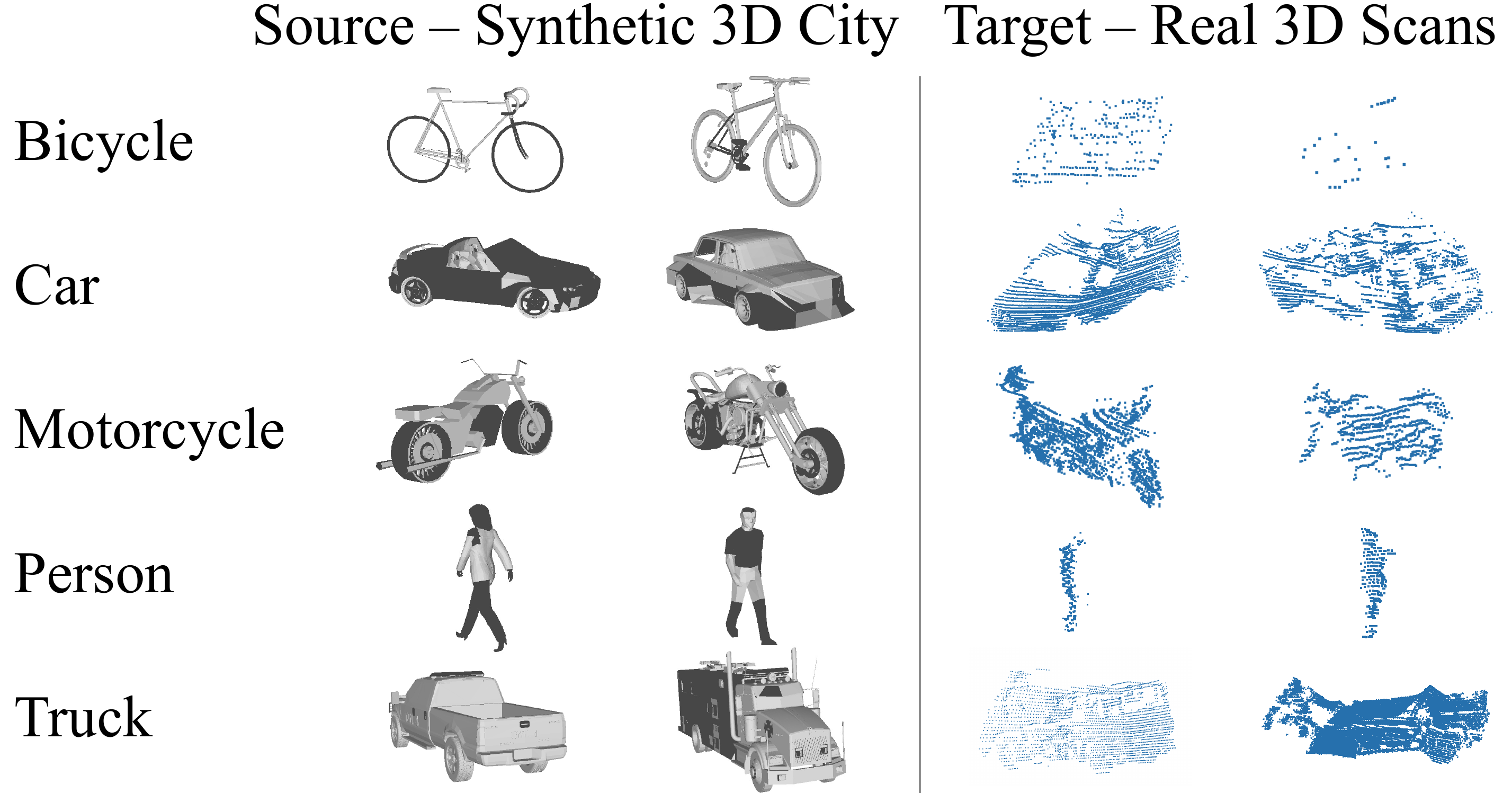}
    \caption{Examples of synthetic and real data in the Point-Syn2Real dataset for the outdoor setting.}
    \label{fig:dataset_vis}
\end{figure}

For the outdoor setting, we obtained five representative categories from 3D Warehouse\footnote{\url{https://3dwarehouse.sketchup.com/}}, ShapeNet, and SemanticKITTI~\cite{dataset_semkitti} datasets. 
We collected CAD models from 3D Warehouse and ShapeNet to construct the synthetic source domain, which we call the 3D\_City subset. We take the real LiDAR object scans from SemnaticKITTI as the real-world target domain.
As the outdoor 3D object scans in SemnaticKITTI are annotated point clouds, we only selected objects that have at least 30 points to make up the SemKitti\_Obj subset. In Figure~\ref{fig:dataset_vis}, we show some examples of synthetic and real data. 
The class distribution of the dataset in the outdoor setting is highly imbalanced. It reflects the availability of public 3D models as well as the distribution of the available real-world data mostly collected for autonomous driving applications.

\section{Evaluation} \label{sec:eval}



We use DGCNN~\cite{dgcnn} with the neighborhood size of $k=20$ as the point cloud encoder.
For training, we use a batch size of 32 and a maximum epoch number of 80.
We use the Adam optimization algorithm with the learning rate set to $0.001$ and the weight decay to $5\times 10^{-5}$. The values of other hyperparameters can be found in the provided code. 
We implement model training and evaluation using PyTorch~\cite{pytorch} and an NVIDIA GTX 3090 GPU. Up to 5GB of GPU memory was used during our experiments.

\begin{table*}[!t]
    \centering
    \footnotesize
    \caption{Performance comparison of the considered approaches in the indoor settings.}
    \begin{tabular}{lccccccc}
    \toprule
    \multirow{2}{*}{Approach}  & \multicolumn{3}{c}{ModelNet to ScanNet} & \multicolumn{3}{c}{ShapeNet to ScanNet} \\
    \cmidrule(lr){2-4} \cmidrule(lr){5-7}  
     {}                 & acc. ($\%$)           & F1 ($\%$)            & MCC                        & acc. ($\%$)         & F1 ($\%$)          & MCC          \\
    \midrule
    Supervised               & 76.37         & 75.85         & \multicolumn{1}{c|}{0.68}  & 76.37       & 75.85       & 0.68         \\
    Baseline                 & 31.09         & 32.98       & \multicolumn{1}{c|}{0.23}      & 24.02       & 28.36       & 0.17         \\
    \midrule
    PointDAN                 & 33.00         & -           & \multicolumn{1}{c|}{-}      & 33.90        & -           & -           \\
    MMD                      & 32.28	     & 35.20       & \multicolumn{1}{c|}{0.22}   & 35.95	    & 37.13      & 	0.23 \\
    DANN                     & 33.75	     & 35.78       & \multicolumn{1}{c|}{0.23}   & 38.33	    & 39.50      &  0.25 \\
    DefRec+PCM               & 51.80         & -           & \multicolumn{1}{c|}{-}      & 54.50        & -          & -     \\
    \midrule
    Point-Syn2Real A         & 51.33	     &  53.02      & \multicolumn{1}{c|}{0.39}      & 50.37	    &  50.70      &  0.37      \\
    Point-Syn2Real A+S       & 57.15	     &  58.96      & \multicolumn{1}{c|}{0.46}      & 58.90	    &  60.11      &  0.48     \\
    \textbf{Point-Syn2Real A+S+E}     & \textbf{59.13}   & 59.30       & \multicolumn{1}{c|}{\textbf{0.48}}      & \textbf{63.48}   & \textbf{62.47}  & \textbf{0.52}     \\
    \midrule
    MMD+A+S                  & 57.72	     & \textbf{59.52}       & \multicolumn{1}{c|}{0.47}      &  59.98	&  60.78      &  0.48      \\
    DANN+A+S                 & 57.60	     & 58.33       & \multicolumn{1}{c|}{0.45}      &  57.26	&  58.82      &  0.45      \\
    \bottomrule
    \end{tabular}
    \label{tab:indoor_compare}
\end{table*}


In our evaluations, we use common classification metrics including overall accuracy and weighted average F1-score~\cite{pointdan_nips19}. In addition, we calculate the Matthews correlation coefficient (MCC) to measure the performance of multi-class classification, especially given that the considered cross-domain datasets are imbalanced. An MCC value of $+1$ indicates perfect prediction, $0$ no better than random prediction, and $-1$ perfect opposite prediction.
For semi-supervised cross-domain learning, we use labeled synthetic data from the source domain and unlabeled real point clouds from the target domain to train the model.
We use a small set of labeled real point clouds from the target domain to validate the model fit and tune the hyper-parameters.
We evaluate the eventual learned model on a held-out target-domain test set that is unseen during the training.


We compare the performance of the proposed approach with a number of existing baseline and state-of-the-art approaches as listed bellow.
\setlist{nolistsep}
\begin{itemize}[noitemsep]
\item{\textbf{Supervised}: The model trained on labeled real-world data from the target domain. It sets an upper bound on the performance of all cross-domain learning approaches.}
\item{\textbf{Baseline}: The model trained only on the synthetic data of the source domain with no domain adaptation, multi-view simulation, or random augmentation.}
\item{\textbf{PointDAN}~\cite{pointdan_nips19}: A domain-adaptation-based approach that utilizes local geometric structures and the global feature distribution.}
\item{\textbf{MMD}~\cite{mmd}: The maximum mean discrepancy approach that uses a discrepancy loss to align the global features between the source and target domains.}
\item{\textbf{DANN}~\cite{dann}: The domain adversarial neural networks approach that utilizes adversarial training to align the global features across the source and target domains.}
\item{\textbf{DefRec+PCM}~\cite{defrec_pointda_wacv19}: A state-of-the-art approach that performs self-supervised deformation-reconstruction (DefRec) to learn cross-domain features using the point cloud mixup (PCM) procedure.}

\item{\textbf{Point-Syn2Real}: The proposed approach.}

\end{itemize}

We denote the multiview point cloud simulation described in section~\ref{mvs} by S, the data augmentation described in section~\ref{da} by A, and the inclusion of entropy loss for semi-supervised learning by E. 

\subsection{Indoor Object Classification}

In a typical indoor setting, common objects are furniture such as table, chair, and bookshelf/cupboard. We collect ten different types of furniture 3D models to train the point cloud encoder and extract key features of these objects. 
In indoor settings, 3D LiDAR scans usually have better resolution and lower noise compared with outdoor settings. As indoor areas are often smaller than outdoor areas of interest, it is easier to obtain high-resolution scans. In addition, indoor settings are generally more controlled and stable. Therefore, the scans are less likely to be contaminated with high levels of noise.
Nonetheless, recognizing objects in the indoor scans, e.g., ScanNet, can be challenging as the object scans are often partially occluded. Especially, for some objects such as a bathtub and a bookshelf, only their top or forward facets are scanned due to the nature of their usage.


The results presented in Table~\ref{tab:indoor_compare} for the ModelNet to ScanNet case show that when augmented via random rotations and additive Gaussian noise, the proposed approach outperforms the earlier domain adaptation method PointDAN, which aligns the distribution of the features learned in the source and target domains. There is a similar observation for the ShapeNet to ScanNet case where the accuracy is improved from 33.90\% (for PointDAN) to 50.37\%.

However, the augmentation alone does not represent the real world, since objects may face different directions in the point cloud coordinate system when they are scanned in the real world.
To account for this, we generate simulated training data from multiple view points which results in slightly varied samples of the same object. 
Benefiting from both augmentation (A) and multiview simulation (S), Point-Syn2Real A+S, further improves the performances. We incorporate the knowledge of the unlabeled target domain data into the training to further regularize the model using the entropy loss for the unlabeled data (E) and adapt it to the target domain. The full model, Point-Syn2Real A+S+E,
outperforms the state-of-the-art approach DefRec+PCM by 7.33\% and 8.98\% in the ModelNet to ScanNet and ShapeNet to ScanNet cases, respectively. MCC score is also improved significantly compared to all the existing methods. Overall, it is evident that the proposed approach offers significant performance improvement in the indoor settings. In addition, our experiments demonstrate that our semi-supervised learning approach through the use of information entropy loss for unlabeled data outperforms more complex domain adaptation methods. The detailed ablation study and comparison with the existing domain-adaptation-based methods, e.g., MMD and DANN, are provided in Sections~\ref{exp:ablation} and \ref{exp:da_methods}.

\subsection{Outdoor Object Classification}

For evaluation on outdoor objects, we extract the real object scans from the SemanticKitti~\cite{dataset_semkitti} autonomous driving dataset. The LiDAR point cloud data in this dataset has been collected using a fast-moving vehicle while the scanned objects themselves may also be moving. In addition, the scanner and the objects are relatively distant. Compared to indoor settings, outdoor settings are generally more dynamic and larger and the scans are more susceptible to noise and error.
We train the model on labeled synthetic source domain, i.e. 3D\_City, and adapt the model to target domain during the training with unlabeled real point cloud from SemKitti\_Obj training split. A held-out test split from target domain is used for testing.
In Table~\ref{tab:outdoor_compare}, we present the performance evaluation results for the considered outdoor setting. 

Both Point-Syn2Real~A and Point-Syn2Real~A+S perform better than the Baseline approach attesting to the effectiveness of the utilized random augmentation and multiview simulation. MMD has high accuracy and F-1 score, close to those of the Supervised approach. However, its MCC value is substantially lower than that of \enquote{Supervised}. This is mainly because MMD is able to classify the more common classes with good accuracy but it fails with the classes that have low frequency.
The proposed Point-Syn2Real~A+S+E approach offers significant improvements over other considered approaches in terms of all three metrics and draws close to the upper bounds set by the Supervised approach. This means that the combination of random augmentation, multiview simulation, and semi-supervised learning appreciably enhances the ability of cross-domain object classification models to generalize to the target domain with minimal supervision.

\begin{table}[!htb]
\centering
\footnotesize
\caption{Performance comparison of the considered approaches in an outdoor setting.}
\begin{tabular}{lcccc}
\toprule
\multirow{2}{*}{Approach}   & \multicolumn{3}{c}{3D\_City to   SemKitti\_Obj}                            \\
\cmidrule(lr){2-4} 
{}                       & \multicolumn{1}{c}{acc ($\%$)} & \multicolumn{1}{c}{F1 ($\%$)} & \multicolumn{1}{c}{MCC} \\
\midrule
Supervised     & 97.06                   & 96.45                  & 0.67   \\
Baseline       & 19.76                   & 29.61                  & 0.09                    \\
\midrule
MMD            & 69.31	& 78.89	& 0.15         \\
DANN           & 42.59	& 57.13	& 0.10         \\
\midrule
Point-Syn2Real A        & 44.69	& 58.58	& 0.14         \\
Point-Syn2Real A+S      & 96.00	& 94.85	& 0.50         \\
\textbf{Point-Syn2Real A+S+E}    & \textbf{96.57}	& \textbf{95.39}	& \textbf{0.57}         \\
\midrule
MMD+A+S        & 96.37	& 95.37	& 0.55         \\
DANN+A+S       & 96.02	& 94.92	& 0.51         \\
\bottomrule
\end{tabular}
\label{tab:outdoor_compare}
\end{table}

\subsection{Qualitative Analysis}
\begin{figure}[th]
    \centering
    \includegraphics[width=0.45\textwidth]{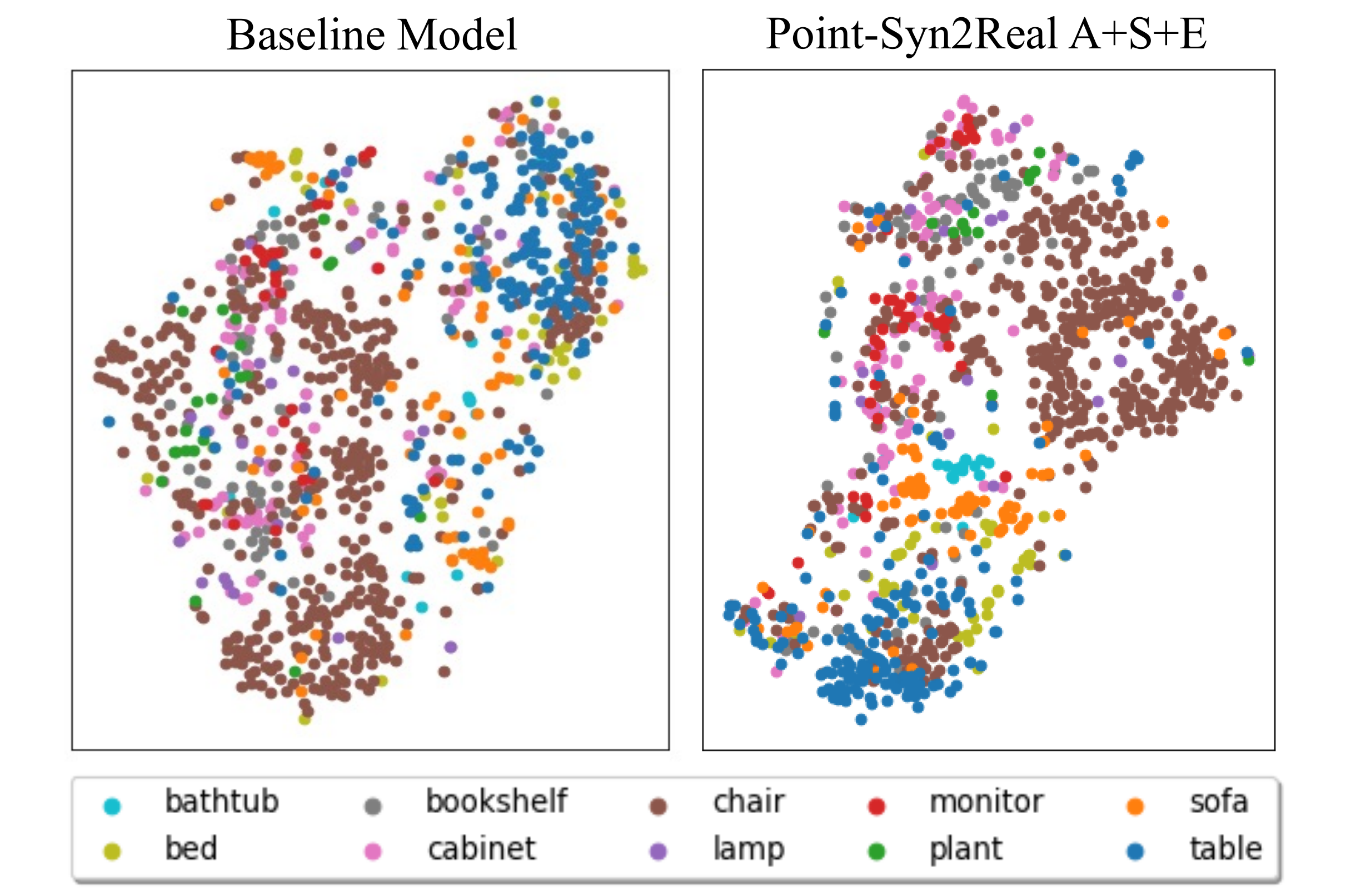}
    \caption{The t-SNE visualizations of the feature space for ModelNet-ScanNet cross-domain learning. The Baseline model is learned from raw synthetic 3D objects, while Point-Syn2Real A+S+E uses more realistic partial scans synthesized from multiple viewpoints. The target domain, ScanNet, is a real-world point cloud dataset. Best viewed in color.}
    \label{fig:vis_tsne}
\end{figure}

\begin{table*}[t]
\centering
\footnotesize
\caption{Performance evaluation results when using DGCNN or PointNet++ as the point cloud encoder (feature extractor).}
\begin{tabular}{lcccccccccc}
\toprule
\multirow{2}{*}{Encoder}                            & \multirow{2}{*}{Approach}  & \multicolumn{3}{c}{ModelNet to ScanNet} & \multicolumn{3}{c}{ShapeNet to ScanNet} & \multicolumn{3}{c}{3D\_City to   SemKitti\_Obj} \\
\cmidrule(lr){3-5} \cmidrule(lr){6-8} \cmidrule(lr){9-11} 
    {}                 & {}                   & acc. ($\%$)          & F1 ($\%$)          & MCC        & acc. ($\%$)          & F1 ($\%$)          & MCC        & acc. ($\%$)            & F1 ($\%$)             & MCC           \\
\midrule
\multirow{2}{*}{DGCNN}      & Baseline & 31.09        & 32.98       & \multicolumn{1}{c|}{0.23}       & 24.02        & 28.36       & \multicolumn{1}{c|}{0.17}       & 19.76            & 29.61            & 0.09 \\
                            & Point-Syn2Real          & \textbf{59.13}        & \textbf{59.30}       & \multicolumn{1}{c|}{\textbf{0.48}}       &   \textbf{63.48}      &  \textbf{62.47}       & \multicolumn{1}{c|}{\textbf{0.52}}       &   \textbf{96.30}         & \textbf{95.06}            & \textbf{0.52}          \\
                           \midrule
\multirow{2}{*}{PointNet++} & Baseline                &  41.61  &  41.58  &  \multicolumn{1}{c|}{0.24}   & 32.17   &  35.23   & \multicolumn{1}{c|}{0.20}   &  65.33   &  76.15            & 0.15           \\
                            & Point-Syn2Real          &  \textbf{60.49}  &  \textbf{56.63}  &  \multicolumn{1}{c|}{\textbf{0.44}}   & \textbf{59.19}  &  \textbf{55.06}  & \multicolumn{1}{c|}{\textbf{0.43}}        & \textbf{95.65}  &  \textbf{94.74}  &  \textbf{0.45}\\
                            \bottomrule
\end{tabular}
\label{tab:applicability}
\end{table*}

Figure~\ref{fig:vis_tsne} illustrates feature space visualizations for ModelNet-ScanNet cross-domain learning using the Baseline approach and the proposed Point-Syn2Real approach produced by the t-SNE~\cite{tsne} algorithm. 
For the visualizations, we randomly select 1000 point clouds from the target domain dataset, i.e., ScanNet, and compute the corresponding object features using the DGCNN encoder trained on the ModelNet synthetic dataset via Baseline or Point-Syn2Real. Each point in Figure~\ref{fig:vis_tsne} represents an object point cloud and is colored according to the corresponding object label.
As seen in the figure, the proposed approach results in features that cluster more distinctly for each object class compared with those of the Baseline approach. This is particularly noticeable for classes that are less prevalent such as ``sofa'' (orange), ``cabinet''(pink), and ``bed'' (yellow). 

\subsection{Choice of Backbone Point Cloud Encoder} \label{sec:applicability}

In Table~\ref{tab:applicability}, we give the performance evaluation results for the considered cross-domain learning settings using two different point cloud feature extraction models, namely, DGCNN and PointNet++.
We implement the Baseline and Point-Syn2Real approaches in the same manner as described in Section~\ref{sec:eval}.
The results in Table~\ref{tab:applicability} show that both encoders lead to similar performance.
Overall, Point-Syn2Real can benefit from both considered point cloud encoders, although, in general, DGCNN is slightly more advantageous hence is our primary choice.


\subsection{Ablation Study} \label{exp:ablation}

We conduct an ablation study to better understand the relative contribution of each component in the proposed Point-Syn2Real approach. In Tables~\ref{tab:indoor_compare} and \ref{tab:outdoor_compare}, we examine the benefits of including augmentation (A), multiview simulation (S), and entropy loss (E) in both indoor and outdoor settings.
For the considered indoor settings, as shown in Table~\ref{tab:indoor_compare}, including the random augmentation alone increases the accuracy significantly, i.e., from 31.09\% to 51.33\% in the ModelNet to ScanNet case and from 24.02\% to 50.37\% in the ShapeNet to ScanNet case, compared to Baseline. This suggests that augmentation is an effective way of enhancing the generalization capacity with relatively small training datasets such as ModelNet. For the considered outdoor setting, as seen in Table~\ref{tab:outdoor_compare}, augmentation alone does not lead to good performance while, together with multiview simulation, it can improve the performance significantly. 
Semi-supervised learning through the use of the entropy loss improves the performance in both indoor and outdoor settings. Especially, in the considered outdoor setting, it increases the MCC value from 0.50 to 0.57. Minimizing the entropy of the posterior class probabilities predicted by the classifier for the unlabeled target training data encourages the classifier to make more confident predictions. This helps the learned model better generalize to unseen data from the target domain. The inclusion of the entropy loss can also be perceived as a form of regularization that prevents the learned model from overfitting to the source domain without relying on any labeled data from the target domain.

\begin{figure}[t]
    \centering
    \includegraphics[width=0.45\textwidth]{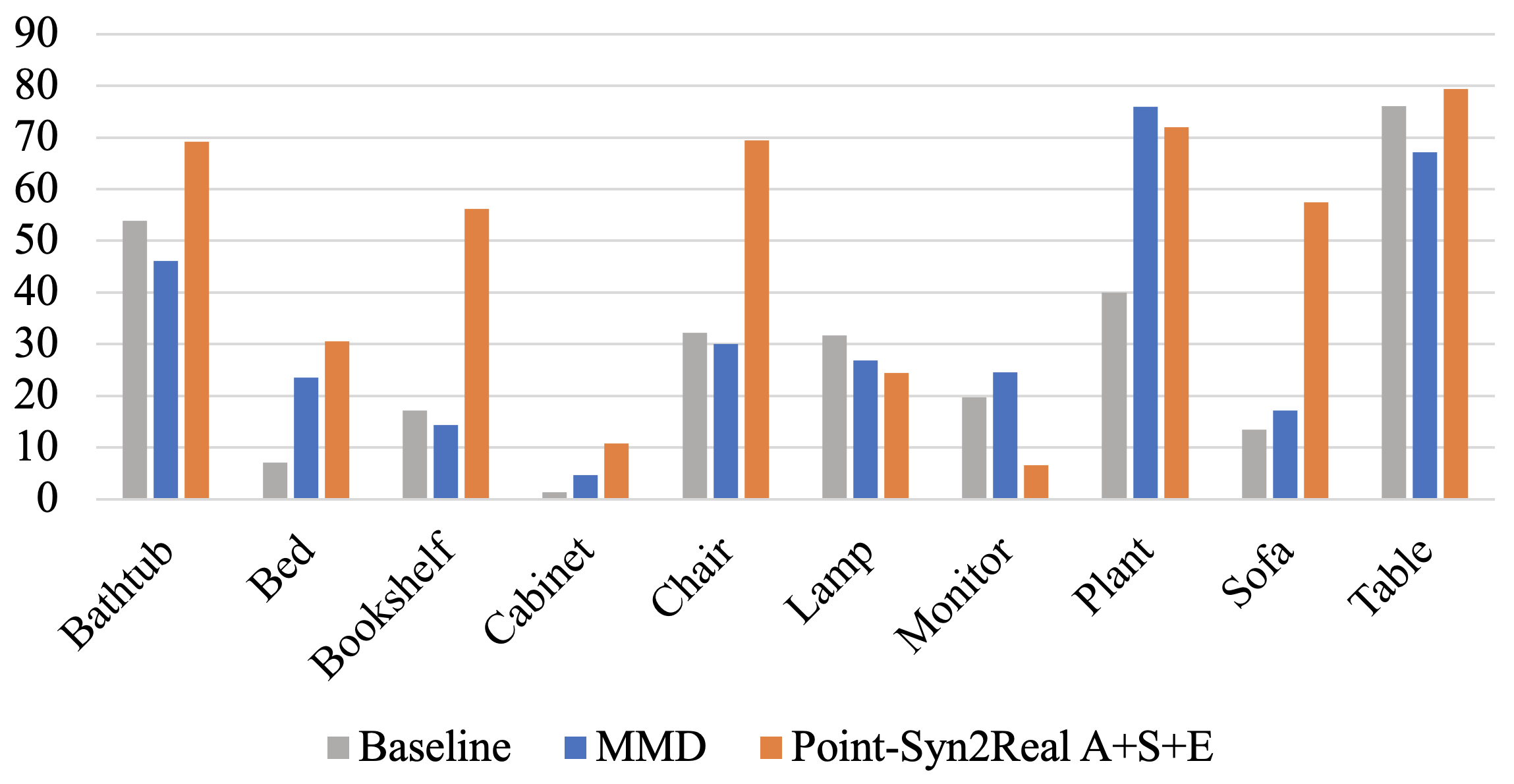}
    \caption{Class-wise accuracy values ($\%$) for the ModelNet to ScanNet case.}
    \label{fig:exp_class_acc}
\end{figure}

Figure~\ref{fig:exp_class_acc} shows the class-wise accuracy of the Baseline, MMD, and Point-Syn2Real A+S+E approaches for the ModelNet to ScanNet case. The results indicate that Point-Syn2Real has the best accuracy for most classes. Especially, the accuracy for the Chair class is about 70\% with Point-Syn2Real while it is around 30\% with Baseline and MMD. The accuracy of Point-Syn2Real is lower than that of MMD for only three classes of Lamp, Monitor, and Plant. It is also interesting to observe that MMD is less accurate than Baseline for five classes. In general, there appears ample room for further improvement considering the class-wise accuracy values, although our proposed approach achieves appreciable improvement over the state-of-the-art.

\subsection{Discussion on Domain Adaptation} \label{exp:da_methods}


In developing Point-Syn2Real, we aim at learning models from synthetic data (source domain) that can generalize to corresponding real-world data (target domain) via simulating the data collection, augmenting the synthesized data, and exploiting the information available through unlabeled target domain data.
Nonetheless, methods based on domain adaptation (DA) such as MMD and DANN have demonstrated promising results in similar tasks pertaining to 2D computer vision. MMD calculates discrepancy and DANN applies adversarial training to adapt the distribution of the global features learned from the source domain to those of the target domain.
Tables~\ref{tab:indoor_compare} and \ref{tab:outdoor_compare} include the performance evaluation results for the mentioned DA-based approaches as well. 
The results show that the considered DA-based approaches alone do not offer any significant benefit. They rather imply that DA for 3D point cloud data is a challenging research question.
We conduct further experiments by applying random augmentation and multiview simulation in conjunction with the DA-based approaches, results of which are indicated by MMD+A+S and DANN+A+S. The results show that the considered data augmentation and multiview simulation are not only beneficial on their own right but also essential for achieving good generalizability across synthetic and real point cloud data domains regardless of the approach taken to adapt the domains.




\section{Conclusion} \label{sec:conclusion} 

We introduced a synthetic-to-real semi-supervised cross-domain learning approach, named Point-Syn2Real, to learn 3D point cloud classification models that can generalize from synthetic domain to real world.
Point-Syn2Real produces synthetic object point clouds by simulating their LiDAR scans from multiple viewpoints while inducing partial occlusions that may occur in real-world 3D scans. It then augments the simulated point clouds by applying random rotations and adding Gaussian noise. The synthesized point clouds are used to train the object classifier that includes a suitable point cloud encoder. 
To mitigate the likelihood of overfitting to the synthetic data of the source domain and hence improve the performance, we incorporate the entropy loss associated with the available unlabeled real data from the target domain into the training objective function.
Through extensive experimentation with synthetic and real data in both indoor and outdoor settings, we showed that Point-Syn2Real outperforms several relevant existing state-of-the-art approaches. This is because the utilization of data augmentation, multiview simulation, and entropy loss enables Point-Syn2Real to better generalize the knowledge learned from the synthetic point cloud data to the real-world data.
We also created a new point cloud dataset, Point-Syn2Real, that can be used to evaluated the performance of point cloud synthetic-to-real cross-domain learning methods.

\newpage
{\small
\bibliographystyle{ieee_fullname}
\bibliography{paper_ref}
}

\end{document}


\title{Supplementary Material}

\maketitle
\thispagestyle{empty}

\section{Appendix}

\subsection{Statistics of the Point-Syn2Real indoor dataset.}

\begin{table*}[h]
\centering
\small
\caption{Statistics of the Point-Syn2Real indoor dataset.}
    \begin{tabular}{lrrrrrrrr}
    \toprule
     \multirow{2}{*}{Class} & \multicolumn{2}{c}{Source - ModelNet} & \multicolumn{2}{c}{Source - ShapeNet} & \multicolumn{2}{c}{Target - ScanNet - Train} & \multicolumn{2}{c}{Target - ScanNet - Test} \\
     \cmidrule(lr){2-3} \cmidrule(lr){4-5} \cmidrule(lr){6-7} \cmidrule(lr){8-9}
      {}   & Instances      & Ratio     & Instances   & Ratio  & Instances   & Ratio & Instances   & Ratio \\
    \midrule
    Bathtub   & 106                  & \multicolumn{1}{r|}{2.53\%}         & 599               & \multicolumn{1}{r|}{3.45\%}      & 98   & \multicolumn{1}{r|}{1.60\%}       & 26                & 1.47\%           \\
    Bed       & 515                  & \multicolumn{1}{r|}{12.31\%}        & 167               & \multicolumn{1}{r|}{0.96\%}      & 329    & \multicolumn{1}{r|}{5.38\%}       & 85                & 4.80\%           \\
    Bookshelf & 572                  & \multicolumn{1}{r|}{13.67\%}        & 310               & \multicolumn{1}{r|}{1.78\%}      & 464    & \multicolumn{1}{r|}{7.59\%}       & 146               & 8.25\%           \\
    Cabinet   & 200                  & \multicolumn{1}{r|}{4.78\%}         & 1,076              & \multicolumn{1}{r|}{6.19\%}      & 650    & \multicolumn{1}{r|}{10.64\%}       & 149               & 8.42\%           \\
    Chair     & 889                  & \multicolumn{1}{r|}{21.25\%}        & 4,612              & \multicolumn{1}{r|}{26.54\%}     & 2,578    & \multicolumn{1}{r|}{42.19\%}       & 801               & 45.28\%          \\
    Lamp      & 124                  & \multicolumn{1}{r|}{2.96\%}         & 1,620              & \multicolumn{1}{r|}{9.32\%}      & 161    & \multicolumn{1}{r|}{2.64\%}       & 41                & 2.32\%           \\
    Monitor   & 465                  & \multicolumn{1}{r|}{11.12\%}        & 762               & \multicolumn{1}{r|}{4.38\%}      & 210   & \multicolumn{1}{r|}{3.44\%}       & 61                & 3.45\%           \\
    Plant     & 240                  & \multicolumn{1}{r|}{5.74\%}         & 158               & \multicolumn{1}{r|}{0.91\%}      & 88    & \multicolumn{1}{r|}{1.44\%}       & 25                & 1.41\%           \\
    Sofa      & 680                  & \multicolumn{1}{r|}{16.26\%}        & 2,198              & \multicolumn{1}{r|}{12.65\%}     & 495   & \multicolumn{1}{r|}{8.10\%}       & 134               & 7.57\%           \\
    Table     & 392                  & \multicolumn{1}{r|}{9.37\%}         & 5,876              & \multicolumn{1}{r|}{33.81\%}     & 1,037   & \multicolumn{1}{r|}{16.97\%}       & 301               & 17.02\%          \\
    \midrule
    Total     & 4,183                 & \multicolumn{1}{r|}{100.00\%}          & 17,378             & \multicolumn{1}{r|}{100\%}       & 6,110   & \multicolumn{1}{r|}{100.00\%}       & 1769              & 100.00\%        \\
    \bottomrule
    \end{tabular}
    \label{tab:dataset_indoor}
\end{table*}

\subsection{Statistics of the Point-Syn2Real outdoor dataset.}

\begin{table}[h]
\centering
\small
\caption{Statistics of the point-Syn2Real outdoor dataset.}
    \begin{tabular}{lrrrrrr}
    \toprule
    \multirow{2}{*}{Class}           & \multicolumn{2}{c}{Source - 3D\_City}  & \multicolumn{2}{c}{Target - SemKitti\_Obj - Train} & \multicolumn{2}{c}{Target - SemKitti\_Obj - Test} \\
    \cmidrule(lr){2-3} \cmidrule(lr){4-5} \cmidrule(lr){6-7} 
    {}         & Instances      & Ratio        & Instances      & Ratio           & Instances    & Ratio        \\
    \midrule
    Bicycle    & 30                   & \multicolumn{1}{r|}{0.96\%}    & 272 & \multicolumn{1}{r|}{0.45\%}   & 261                    & 1.54\%            \\
    Car        & 2,458                 & \multicolumn{1}{r|}{78.88\%}   & 58,035 & \multicolumn{1}{r|}{95.25\%}    & 16,140                  & 95.05\%           \\
    Motorcycle & 235                  & \multicolumn{1}{r|}{7.54\%}    & 814 & \multicolumn{1}{r|}{1.34\%}    & 332                    & 1.96\%           \\
    Person     & 30                   & \multicolumn{1}{r|}{0.96\%}    & 411 & \multicolumn{1}{r|}{0.67\%}    & 69                     & 0.41\%           \\
    Truck      & 363                  & \multicolumn{1}{r|}{11.65\%}   & 1,395 & \multicolumn{1}{r|}{2.29\%}    & 179                    & 1.05\%           \\
    \midrule
    Total      & 3,116                 & \multicolumn{1}{r|}{100.00\%}     & 60,927 & \multicolumn{1}{r|}{100.00\%}     & 16,981                  & 100.00\%           \\
    \bottomrule
    \end{tabular}
    \label{tab:dataset_outdoor}
\end{table}

\subsection{Related work on other cross-domain learning methods}

There are other cross-domain learning methods that tackle different tasks and are less relevant to our work.
Among them, xMUDA~\cite{xmuda} leverages multiple modalities to adapt the model learned from a day-time source domain to a night-time target domain. Cross-sensor~\cite{cross-sensor} learns shared representations for point cloud data collected from different types of LiDAR sensors. VehicleBEV~\cite{birdview} learns from synthetic data to infer on real LiDAR point cloud data. However, instead of learning directly from point cloud data, it converts the problem to a 2D image domain adaptation problem by projecting 3D point clouds onto 2D bird's eye view (BEV) images.

\subsection{Simulation User Interface}

\begin{figure}[h]
    \centering
    \includegraphics[width=0.8\textwidth]{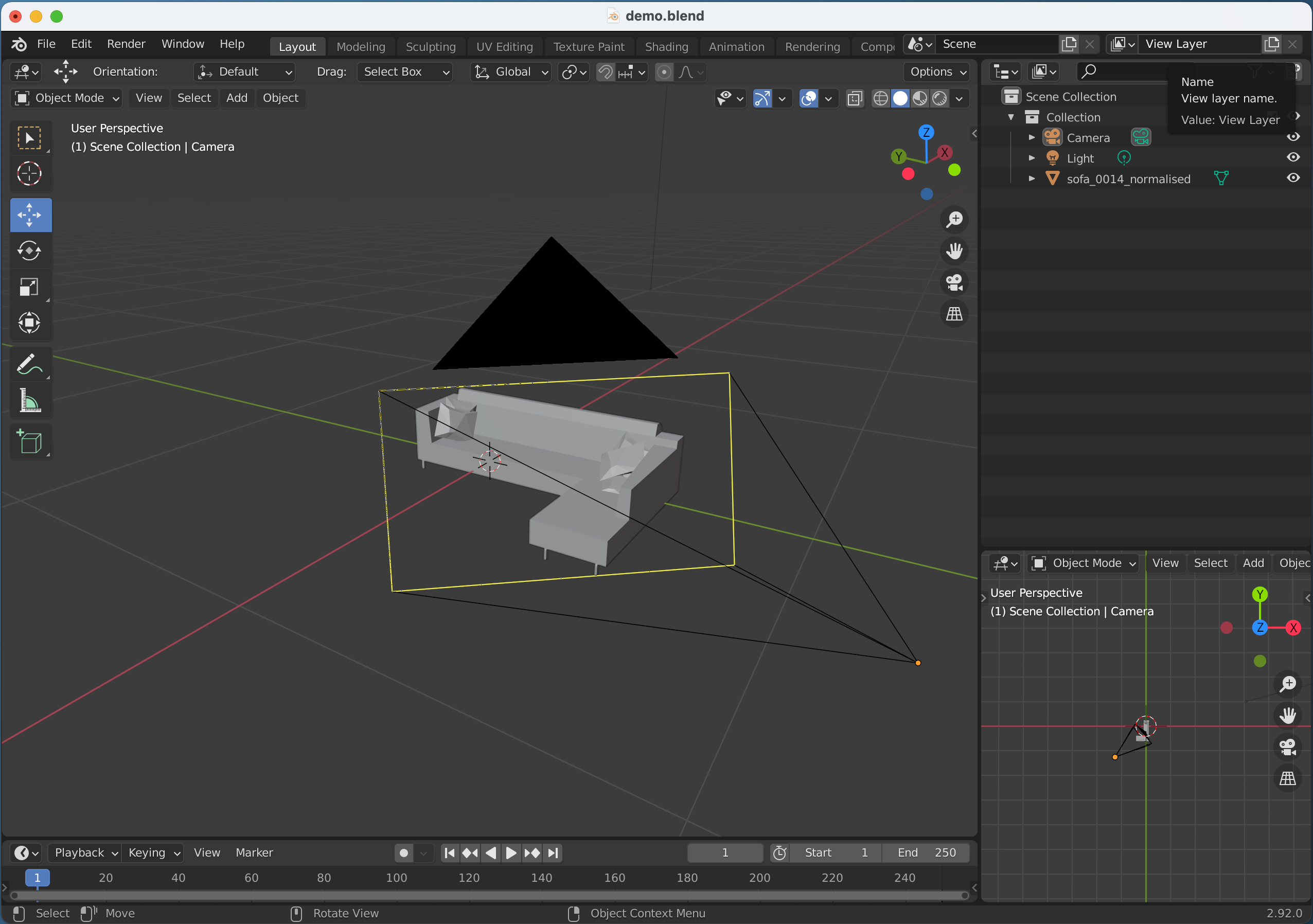}
    \caption{Blender User Interface for multiview simulation.}
    \label{fig:blender_ui}
\end{figure}
{\small
\bibliographystyle{ieee_fullname}
\bibliography{paper_ref}
}